\documentclass[11pt]{article}
\usepackage[preprint]{acl}

\usepackage{times}
\usepackage{latexsym}

\usepackage[T1]{fontenc}

\usepackage[utf8]{inputenc}

\usepackage{microtype}

\usepackage{inconsolata}

\usepackage{graphicx}
\usepackage{svg}
\usepackage{amsmath}
\usepackage{booktabs}
\usepackage{multirow}
\usepackage{rotating}
\usepackage{tipa}

\usepackage{tikz}
\newcommand{\tstroke}{%
  \tikz[baseline=(X.base)]{
    \node[inner sep=0pt] (X) {t};
    \draw[line width=0.03em] (X.west)++(0.01em,0.02ex) -- ++(0.26em,0);
  }%
}

\title{Pretraining Finnish ModernBERTs}

\author{
  \textbf{Akseli Reunamo\textsuperscript{1}},
  \textbf{Laura-Maria Peltonen\textsuperscript{2,3}},
  \textbf{Hans Moen\textsuperscript{4}},
  \textbf{Sampo Pyysalo\textsuperscript{1}},
\\
\\
\textsuperscript{1}University of Turku, Department of Computing, 
\\
\textsuperscript{2}University of Eastern Finland and Kuopio University Hospital,
\\
\textsuperscript{3}University of Turku and Turku University Hospital,
\\
\textsuperscript{4}Aalto University, Department of Computer Science,
\\
\\
  \small{
    \textbf{Correspondence:} \href{mailto:akseli.y.reunamo@utu.fi}{akseli.y.reunamo@utu.fi}
  }
}

\begin{document}
\maketitle
\begin{abstract}
This paper reports on pretraining ModernBERT encoder models in six different sizes, ranging from 51M to 475M parameters, with a focus on limited multilingualism, emphasizing languages relevant to Finland. Our models are competitive with, or superior to, existing multilingual models. They outperform monolingual models on tasks that require a context longer than 512 tokens. We present empirical results on using different data in the final stage of training. The \href{https://github.com/rakseli/finnish-modernberts}{code} and \href{https://huggingface.co/collections/TurkuNLP/finnish-modernberts-685bb5f2ab4d39d6480a16d4}{models} are publicly released.
\end{abstract}

\section{Introduction}
Since the introduction of Bidirectional Encoder Representations from
Transformers (BERT) \cite{bert}, encoder-only language models based on the transformer architecture \cite{transformer}, have been popular for natural language understanding (NLU) tasks such as retrieval, classification, and ranking \cite{debertav3,roberta}. These models are competitive with orders of magnitude larger decoder-only models in NLU tasks \cite{gpt,gpt2,llama31,euroeval}. The small inference and fine-tuning memory requirements of encoder-only models\footnote{Computation cost in autoregressive decoding is rarely a limiting factor in classification tasks, as often only a single label needs to be decoded. However, the processor's memory is the fundamental limitation \protect{\cite{megatron-lm}}.} are appealing characteristics that promote the adoption of modeling-based solutions in large-scale and resource-constrained applications.

XLM-R \cite{xlmr} has been a favored choice for multilingual NLU tasks since its release. Most recently, new multilingual encoder models, such as mmBERT \cite{mmbert} and EuroBERT \cite{eurobert}, have demonstrated competitive or superior performance on multilingual benchmarks. However, none of the multilingual models have broken "the curse of multilinguality" \cite{xlmr}, and perform disadvantageously compared to their monolingual equivalents.

While common multilingual models, such as mBERT and XLM-R, have been trained with Finnish in their training data, they struggle with the language's rich morphology, where single words may take thousands of forms through as many as 15 grammatical cases, often tokenize Finnish text inefficiently, and can not handle sequences longer than 512 tokens. As Finnish often constitutes only a small part of multilingual training data, this results in suboptimal performance compared to well-represented languages. Dedicated Finnish language models \cite{finbert} address these limitations, achieving better performance with fewer parameters. However, a regionally relevant model managing Finland's multilingual context, where code-switching between Finnish, Swedish, and English is common, has not yet been introduced.

In this research, we present a collection of encoder-only models based on the ModernBERT architecture \cite{modernbert}, trained on 362 to 448 billion tokens in languages used in Finland\footnote{Finnish and Swedish are the national languages of Finland. Multiple Sámi languages are spoken in Finland, with Northern Sámi being the most widely spoken. English is not an official language of Finland, but it is commonly used. We included Latin for possible clinical use.}: Finnish, Swedish, Northern Sámi, English, and Latin. We aimed to train regionally relevant models that can handle sequences longer than 512 tokens, with an emphasis on Finnish, and limited multilinguality. The collection comprises six models, each with varying sizes, context lengths, and vocabulary sizes. We make the models\footnote{\url{https://huggingface.co/collections/TurkuNLP/finnish-modernberts-685bb5f2ab4d39d6480a16d4}} and accompanying code\footnote{\url{https://github.com/rakseli/finnish-modernberts}} publicly available under a permissive license. We refer to the models we trained as Finnish ModernBERTs. We additionally release checkpoints that can be used in continued pretraining.  We measure the effect of using different data in the late stages of training in a multilingual setting. The models demonstrate competitive performance compared to other multilingual and monolingual models in multiple NLU tasks, and outperform other multilingual models in long context out-of-domain retrieval.

\section{Pretraining}

\subsection{Hardware}
We did all processing and development on the LUMI supercomputer\footnote{\url{https://www.lumi-supercomputer.eu/}}. The main LUMI GPU partition comprises 2978 nodes, each equipped with four AMD MI250x GPUs. Each MI250x contains two GPU dies. Additionally, LUMI has CPU nodes equipped with main memory up to four TiB. We trained the models using eight nodes, which allows for training with 64 distributed processes. 

\subsection{Software}
We trained the models using custom code that uses modules from Transformers (version 4.51.1) \cite{hugginface-transformers} and PyTorch (version 2.6.0) \cite{pytorch}. We utilize PyTorch torch.nn.parallel.DistributedDataParallel and torch.distributed.optim.ZeroRedundancyOptimizer to scale training for larger batch sizes. We used Flash Attention 2 (version 2.7.3) as the attention backend \cite{flashattention2}.

\subsection{Tokenization}
For each model size, we trained a BPE tokenizer with an optimal vocabulary estimated based on the available Floating-point Operations budget. For the optimal vocabulary size (Table \ref{tab:vocabsize}), the training data budget was set to a fixed 400B tokens, and we used the Parametric Fit of Loss Formula \cite{vocabscalinglaws} for the prediction. For computational efficiency, each vocabulary size was increased to be divisible by 64. In addition to model-specific tokenizers, one baseline BPE tokenizer with a fixed vocabulary size of 128K was trained. Each tokenizer was trained on 48GB of text sampled from the training corpus. 

\begin{figure*}
  \centering
  \includegraphics[width=\linewidth]{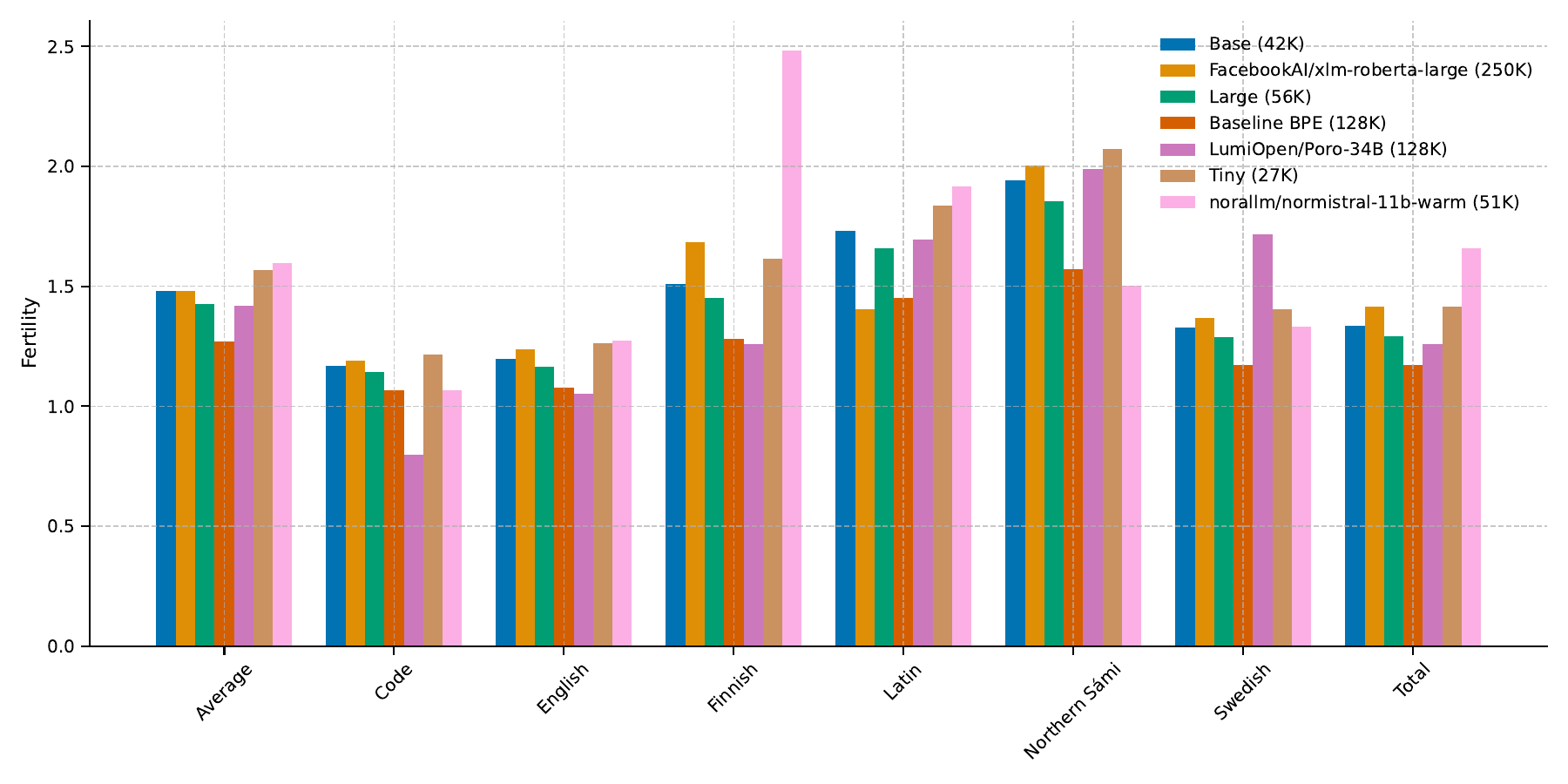}
  \caption{Larger vocabularies predictably have better fertilities compared to smaller ones.}
  \label{fig:fert}
\end{figure*}

We utilized the special token set from the original BERT \cite{bert} and adapted the tokens for repeated spaces from GPT-NeoX-20B \cite{gptneox20b}, which are also employed in Olmo \cite{olmo} and ModernBERT \cite{modernbert}. We evaluated the tokenizers' fertilities on the validation set of the pretraining corpus. Figure \ref{fig:fert} shows the differences in the fertilities of the tokenizers among the selected reference tokenizers. The newly trained tokenizers yielded fertilities comparable to those of other multilingual tokenizers. 
\begin{table}[ht]
    \centering
    \begin{tabular}{lll}
        \textbf{Tokenizer}  & \textbf{Predicted optimal} & \textbf{True size} \\
        \hline
        tiny & 27,224 & 27,264 \\ 
        base  & 42,200 & 42,240 \\ 
        large  & 55,571 & 55,616 \\ 
        baseline  & - & 128,000 \\
    \hline
    \end{tabular}
    \caption{Tokenizer vocabulary sizes}
    \label{tab:vocabsize}
\end{table}

\subsection{Architecture}
All models are encoder models based on the ModernBERT architecture \cite{modernbert}, and we follow most of their design choices. Deviating from the ModernBERT, we use Rotary Position Embedding (RoPE) \cite{rope}, $\theta_{i}=base^{-2i/d}$, with rotary base of $base=1000000$ instead of the original 160K used in global attention layers. We chose this rotary base because it can, in theory, extrapolate to a longer context than the 160K rotary base \cite{ropescalinglaws}. The number of parameters is dependent on the number of encoder layers and the vocabulary size. We refer to models as "tiny", "base", and "large". Model names without a suffix use an optimal vocabulary size and were trained with a context length of 16K, and models suffixed by "short" used a vocabulary size of 128K and were trained with a context length of 8192. Details of each model are presented in Table \ref{tab:parameters}. 

\begin{table}[ht]
    \centering
    \begin{tabular}{llllll}
        \textbf{Model}  & \textbf{P} & \textbf{L} & \textbf{H} & \textbf{I} & \textbf{AH}\\
        \hline
        tiny  & 51M        & 6  & 768  & 1152 & 12 \\ 
        tiny-short & 129M  & 6  & 768  & 1152 & 12 \\ 
        base  & 143M       & 22 & 768  & 1152 & 12 \\
        base-short & 209M  & 22 & 768  & 1152 & 12 \\ 
        large       & 401M & 28 & 1024 & 2624 & 16 \\
        large-short & 475M & 28 & 1024 & 2624 & 16 \\
    \hline
    \end{tabular}
    \caption{Model design. The column "P" refers to the number of parameters in millions, "L" refers to the number of encoder layers, "H" refers to the encoder layer dimensionality, "I" refers to the intermediate fully connected feed-forward network dimensionality, and "AH" refers to the number of attention heads.}
    \label{tab:parameters}
\end{table}

\subsection{Training}

Following previous work \cite{modernbert,mosaicbert}, we use the masked language modeling procedure with a 30\% masking rate without the next-sentence prediction \cite{bert}. We use AdamW optimizer \cite{adamw} and a trapezoidal learning rate (LR) scheduler \cite{wsd1,wsd2}. This scheduler is also known as the Warmup-Stable-Decay (WSD) schedule. We follow \citet{modernbert} and use $1-\sqrt{LR}$ in the LR decay phase, and we use batch size warm-up. All models were trained with the same number of optimization steps (138000), but the number of tokens varied due to different global batch sizes, device batch size variation caused by unpadding, and the number of gradient accumulation steps. The difference in input tokens was maximally 19\% (tiny and tiny-short-edu), and 50\% of the models were trained on an equal number of tokens.  We describe the training as a three-step process where we assume that models' parameters are first optimized for short context representations of the tokens (stable phase), then refine the token representations for longer dependencies (context extension phase), and finally, reinforce the representations for the inputs that we believe the model will be used for in the future (annealing phase). 
Training hyperparameters and additional details are provided in Appendix \ref{appendix:a}.

\subsection{Data}
All models are trained with the same pretraining data mixture. The language distribution of the pretraining data used prior to the context extension and annealing is provided in Table \ref{tab:lang_dist}. We used text datasets from diverse sources, including web crawls, news, scientific articles, classical literature, historical texts, Wikipedia, forums, and governmental sources such as policies and laws. Sources underwent various levels of pre-processing, including the removal of low-quality text or boilerplate. We deliberately oversample the datasets considerably (median sampling factor 5, arithmetic mean 11.21, maximum 66 for the Cultura X Latin subset \cite{culturax}) based on experience from previous studies \cite{babylm,debertav3}. Pretraining crawl data, excluding high-quality educational data used for annealing, was globally exactly deduplicated per language \cite{textdedup}, and personally identifiable information (PII) was removed \cite{bloom,multilingual_pii_tool}. Additionally, some individual datasets were PII-removed because we considered them to be high-risk data sources. A complete listing of used datasets and processing details is provided in Appendix \ref{appendix:b}. 

\begin{table}[ht]
    \centering
    \begin{tabular}{lll}
        \textbf{Language}  & \textbf{Tokens (B)} & \textbf{\%} \\ \hline
        Finnish  & 209.09 & 53.6 \\ 
        English  & 80.77 & 20.7 \\ 
        Swedish  & 80.09 & 20.5 \\ 
        Code & 14.12 & 3.6 \\ 
        Cross-lingual  & 3.98 & 1.0 \\
        Northern Sámi & 1.07 & 0.3 \\ 
        Latin  & 0.94 & 0.3 \\
        \hline
        \textbf{Total}  & \textbf{390} & \textbf{100} \\
    \hline
    \end{tabular}
    \caption{Pretraining data language distribution. "B" refers to billions.}
    \label{tab:lang_dist}
\end{table}

\subsubsection{Annealing and context extension data}
\label{sec:annealing}
For the model context length extension, we sampled documents with varying token lengths from the pretraining data, focusing on those longer than the initial context length of 1024 tokens. The resulting data had a more uniform language distribution than the original data. 
For annealing, our aim was to create two high-quality data mixes. The first mix uses context extension data, but we restrict the document sources to those we subjectively deem high-quality based on visual inspection. We refer to this data as "baseline annealing data". In this baseline annealing data, most texts are written in English (over 90\%) and are from the Fineweb-edu fortified dataset \cite{fineweb-edu-fortified,fineweb} (44\%). For the second mix, we used existing classifiers \cite{llama_hplt_edu_classifiers_2025} to predict the educational score for Finnish and Swedish data from HPLT 2.0 cleaned \cite{hpltv2}, and then filtered them based on a score threshold of $\geq2$. We refer to this data as the "edu" data. Then, we complement the edu data with a selected set of sources from the pretraining data mix. The edu data language distribution is more balanced, with most data coming from the Finnish subset of HPLT 2.0 (53\%). Detailed descriptions of both annealing data mixes and context extension data are provided in Appendix \ref{appendix:c}.

\subsection{Environmental, Computational, and Economic costs}

Inspired by the work of \citet{olmo}, we estimated the energy consumption of the model training by using the theoretical maximum power consumption of MI250x GPUs while accounting for energy efficiency (Equation \ref{eq:energy}). 

\begin{equation}
  \label{eq:energy}
    E_{total}=E_{GPU}\cdot N \cdot T \cdot PUE
\end{equation}

where

\begin{tabular}{@{}lp{0.6\linewidth}l@{}}
    \centering
    $E_{total}$ & Total consumed energy \\[0.5ex]
    $E_{GPU}=$ & Theoretical upper bound of energy consumed by the GPU when it utilizes 100\% of its capacity \\[0.5ex]
    $N$ & Number of GPU modules \\[0.5ex]
    $T$ & Training wall time \\[0.5ex]
    $PUE$ & system's Power Usage  Effectiveness \\[0.5ex]
\end{tabular}

To estimate carbon emissions, we multiplied the total power consumption of GPUs by a carbon intensity factor measured in kg CO$_2$ emitted per kWh. Hydroelectric energy sources, which do not involve the construction of a large reservoir, have minimal post-construction emissions \cite{hydropower}. The LUMI supercomputer runs on 100\% renewable run-of-river hydropower energy, thus the carbon intensity factor is set to $0.004$ kg CO$_2$/kWh based on the median run-of-river hydropower emissions \cite{icpp}.

We used the $PUE=1.04$\footnote{reported by LUMI}, $N=32$, $E_{GPU}=560W$\footnote{We used Thermal Design Power reported by \citet{mi250x}}. The total training time, energy consumption, and emissions of our models are represented in Table \ref{tab:emission}. The carbon emissions correspond to driving an average new registered car in the EU for about 900km \cite{averagecar}.

\begin{table}[ht]
    \centering
    \begin{tabular}{llll}
        \textbf{Model}  & \textbf{Wall time (h)} & \textbf{MWh} & \textbf{CO$_2$}\textbf{eq}\\
        \hline 
        tiny  & 80.89 & 1.51 & 6.04 \\ 
        tiny-short & 194.91  & 3.63 & 14.52 \\ 
        base   & 158.66 & 2.96 & 11.84 \\ 
        base-short & 236.60 & 4.41 & 17.64 \\ 
        large   & 286.85 & 5.35 & 21.40\\ 
        large-short & 299.23 & 5.58 & 22.32\\
        \hline
        \textbf{Total} & \textbf{1257.14} & \textbf{23.44} & \textbf{93.76} \\
    \hline
    \end{tabular}
    \caption{Training resource consumption. Variation in training wall times between similar-sized models stems from differences in the number of gradient accumulation steps, the use of chunked cross-entropy in models trained with 128K vocabulary, and failed runs.}
    \label{tab:emission}
\end{table}

A lower financial limit for training a single large model with training optimization in question using a commercial platform with next-generation hardware\footnote{AMD MI300X is used as reference for next-generation hardware and performance increase compared to AMD MI250X is deduced from Peak single precision Matrix performance \protect{\cite{mi300x,mi250x}}. MI250X has a peak of 95.7 TFLOPs and MI300X has peak of 163.4 TFLOPs.} is 9366€\footnote{Average AMD MI300X hourly compute price was 1.67€/h when evaluating several cloud providers in September 2025 and large-short training required 9575.36 GPU hours}

Our estimates are approximate; they do not include other vital components of development, such as debugging and evaluation. We hope these estimates will aid in evaluating the actual cost of model pretraining in the future.

\section{Evaluation}
We evaluated seven versions of our models: models trained with optimal vocabulary and sequence length 16K, vocabulary size 128K and sequence length 8192, optimal vocabulary size and sequence length 1024 (suffix nce), annealing with the edu data (suffix edu) or with the baseline annealing data, and without annealing (suffix cpt). Both annealing data subsets are presented in \ref{sec:annealing}. We compared monolingual and multilingual encoder models\footnote{We select the best-performing encoder-only models for comparison rather than focusing strictly on the number of parameters.} on short and long context tasks.
\subsection{Short context evaluation}
We evaluated the models using a custom version\footnote{\url{https://github.com/rakseli/EuroEval}} of EuroEval \cite{euroeval}, which enabled us to perform a hyperparameter search for optimal learning rates. EuroEval downsamples the original included datasets\footnote{The training splits contain 1024 samples.} permitting more efficient evaluation.

In the original framework, all encoder models are fine-tuned ten times on the training sets with a learning rate of $2e-5$ and an early stopping patience of $2$, and then evaluated on bootstrapped test sets \cite{nielsen-2023-scandeval}. We conducted a hyperparameter search for all models using $learning\ rate \in \{1e-5, 2e-5, 3e-5, 4e-5, 5e-5\}$ and an early stopping patience of $5$. We similarly trained ten models per learning rate and evaluated them using bootstrapped validation splits of the datasets. The final evaluations were conducted using the optimal learning rate identified in the hyperparameter search, and the bootstrapped test splits were used to evaluate the model. The best learning rate was defined as the one that yielded the highest mean F1 score over the task. We evaluated similar-sized monolingual and multilingual models on tasks in Finnish \cite{scandeval,scandisent,turku-ner,tydiqa} (Table \ref{tab:euroeval-fi}), Swedish \cite{scandeval,suc3,swerec} (Table \ref{tab:euroeval-sv}), and English \cite{conll-en,scandeval,squad,sst-5} (Table \ref{tab:euroeval-en}).

\begin{table*}
\centering
\scriptsize
\setlength{\tabcolsep}{3pt}
\begin{tabular}{l l c c c c}
\textbf{Model} & \textbf{N params (M)} & \textbf{scala-fi} & \textbf{scandisent-fi} & \textbf{turku-ner-fi} & \textbf{tydiqa-fi} \\
\cmidrule(lr){3-6} & & \textit{MCC / Ma F1} & \textit{MCC / Ma F1} & \textit{Mi F1 no misc / Mi F1} & \textit{F1 / EM} \\
\midrule
\multicolumn{6}{l}{\textbf{Monolingual}} \\
bert-base-finnish-cased-v1 & 125.17 & 47.16±5.27 / 72.98±2.47 & 90.16±0.50 / 95.08±0.25 & \textbf{82.04±1.33 / 79.35±0.94} & 56.20±1.42 / 35.68±1.82 \\
bert-large-finnish-cased-v1 & 355.20 & \textbf{58.81±2.46 / 78.91±1.23} & \textbf{91.69±0.60 / 95.85±0.30} & 77.57±1.43 / 74.50±1.74 & \textbf{59.91±1.19 / 39.10±1.18} \\
\addlinespace
\midrule
\multicolumn{6}{l}{\textbf{Multilingual}} \\
Finnish-ModernBERT-base & 143.40 & 24.81±6.66 / 61.46±3.62 & 84.59±1.80 / 92.26±0.89 & 56.17±4.80 / 56.03±4.91 & 30.04±1.27 / 14.22±1.25 \\
Finnish-ModernBERT-base-edu & 143.40 & 30.26±4.60 / 64.41±2.50 & 83.71±2.88 / 91.81±1.49 & 61.49±2.25 / 62.16±2.04 & 30.58±1.57 / 14.22±1.73 \\
Finnish-ModernBERT-base-short & 209.35 & 44.04±6.17 / 71.26±3.28 & 80.17±5.76 / 90.03±2.89 & 59.32±2.37 / 60.86±1.84 & 33.67±2.40 / 15.73±1.78 \\
Finnish-ModernBERT-base-short-cpt & 209.35 & 30.81±7.86 / 63.46±5.01 & 81.90±5.44 / 90.91±2.72 & 62.10±3.16 / 62.37±2.67 & 31.40±2.13 / 14.39±1.57 \\
Finnish-ModernBERT-base-short-edu & 209.35 & 34.13±6.76 / 65.80±3.63 & 82.31±2.76 / 91.10±1.39 & 63.29±2.92 / 63.85±2.47 & 29.89±0.98 / 13.06±0.72 \\
Finnish-ModernBERT-large & 401.26 & 51.88±3.07 / 75.39±1.91 & 88.02±2.33 / 93.99±1.18 & 71.11±1.83 / 70.47±1.44 & 43.45±2.92 / 23.47±2.90 \\
Finnish-ModernBERT-large-edu & 401.26 & 52.57±4.67 / 75.38±3.30 & 89.35±0.87 / 94.65±0.44 & 71.43±2.23 / 71.19±1.68 & 45.43±2.19 / 23.83±1.95 \\
Finnish-ModernBERT-large-nce & 401.26 & 49.81±4.13 / 74.58±2.10 & 88.50±2.88 / 94.22±1.47 & 71.16±2.41 / 70.58±2.01 & 42.40±3.43 / 22.17±2.78 \\
Finnish-ModernBERT-large-short & 475.45 & 53.99±4.49 / 76.36±2.68 & 89.48±0.73 / 94.72±0.38 & 72.78±1.41 / 72.22±1.01 & 44.74±4.07 / 23.71±3.43 \\
Finnish-ModernBERT-large-short-cpt & 475.45 & 53.40±4.26 / 76.11±2.38 & 90.21±0.77 / 95.08±0.40 & 72.65±1.97 / 72.29±2.26 & 44.64±3.94 / 23.43±2.58 \\
Finnish-ModernBERT-large-short-edu & 475.45 & \textbf{56.65±3.40 / 77.79±1.87} & 89.93±0.75 / 94.95±0.39 & 71.47±2.34 / 70.85±2.41 & 44.90±3.01 / 23.57±2.31 \\
Finnish-ModernBERT-tiny & 51.65 & 4.94±1.95 / 51.89±1.24 & 76.15±1.93 / 88.05±0.97 & 52.45±1.23 / 53.81±1.05 & 29.63±0.42 / 14.59±0.58 \\
Finnish-ModernBERT-tiny-edu & 51.65 & 3.96±2.40 / 51.72±1.27 & 76.03±1.42 / 87.98±0.73 & 53.13±2.16 / 53.91±2.08 & 30.33±0.64 / 14.61±0.71 \\
Finnish-ModernBERT-tiny-short & 129.11 & 12.05±3.69 / 55.38±2.13 & 79.38±1.81 / 89.66±0.92 & 55.78±2.81 / 56.98±2.94 & 29.57±0.52 / 12.98±0.69 \\
Finnish-ModernBERT-tiny-short-cpt & 129.11 & 10.91±3.30 / 54.96±1.75 & 78.89±1.79 / 89.42±0.91 & 55.44±3.52 / 56.41±3.73 & 30.32±0.57 / 14.11±0.99 \\
Finnish-ModernBERT-tiny-short-edu & 129.11 & 11.52±3.44 / 55.43±1.95 & 79.40±1.79 / 89.67±0.90 & 56.32±2.81 / 57.29±3.00 & 29.55±0.55 / 12.92±0.70 \\
mmBERT-base & 307.53 & 21.49±4.11 / 59.87±2.56 & 80.73±2.19 / 90.32±1.09 & 67.22±3.28 / 64.32±2.89 & 38.44±2.41 / 19.08±1.66 \\
XLM-RoBERTa-large & 561.19 & 50.84±3.76 / 74.32±2.41 & \textbf{90.39±1.12 / 95.18±0.56} & \textbf{84.31±1.35 / 81.93±1.07} & \textbf{56.66±5.70 / 35.34±4.34} \\
\bottomrule
\end{tabular}
\caption{Results for tasks included in EuroEval in the Finnish language. In the metrics header, "MCC" is an abbreviation for Matthews Correlation Coefficient, "Ma" represents macro, "Mi" represents micro, and "EM" represents exact match. The best average scores for monolingual and multilingual models are highlighted in bold.}
\label{tab:euroeval-fi}
\end{table*}

\begin{table*}
\centering
\scriptsize
\setlength{\tabcolsep}{3pt}
\begin{tabular}{l l c c c c}
\textbf{Model} & \textbf{N params (M)} & \textbf{scala-sv} & \textbf{scandiqa-sv} & \textbf{suc3} & \textbf{swerec} \\
\cmidrule(lr){3-6} & & \textit{MCC / Ma F1} & \textit{F1 / EM} & \textit{Mi F1 no misc / Mi F1} & \textit{MCC / Ma F1} \\
\midrule
\multicolumn{6}{l}{\textbf{Monolingual}} \\
roberta-large-1160k & 355.41 & \textbf{76.24±1.30 / 87.74±0.72} & \textbf{53.13±0.86 / 46.76±1.08} & \textbf{79.27±2.28 / 76.65±2.03} & \textbf{77.43±0.65 / 76.11±1.73} \\
\addlinespace
\midrule
\multicolumn{6}{l}{\textbf{Multilingual}} \\
Finnish-ModernBERT-base & 143.40 & 58.79±2.50 / 78.96±1.22 & 29.98±2.03 / 23.35±2.22 & 51.67±3.10 / 53.42±3.09 & 63.10±3.20 / 62.47±4.03 \\
Finnish-ModernBERT-base-edu & 143.40 & 58.81±2.73 / 78.78±1.51 & 29.03±1.22 / 22.02±1.14 & 53.55±2.64 / 55.12±2.74 & 64.58±2.48 / 64.28±2.76 \\
Finnish-ModernBERT-base-short & 209.35 & 62.42±2.15 / 81.07±1.06 & 30.85±1.72 / 24.25±1.75 & 57.57±2.12 / 58.95±1.76 & 66.78±1.97 / 65.71±2.88 \\
Finnish-ModernBERT-base-short-cpt & 209.35 & 59.42±3.91 / 79.36±2.11 & 31.83±1.50 / 25.23±1.89 & 57.52±2.97 / 59.02±2.69 & 65.85±2.52 / 65.64±2.22 \\
Finnish-ModernBERT-base-short-edu & 209.35 & 62.00±3.20 / 80.83±1.72 & 30.81±2.05 / 24.25±2.41 & 57.59±3.00 / 59.24±2.65 & 66.77±2.07 / 65.78±2.75 \\
Finnish-ModernBERT-large & 401.26 & 69.42±3.72 / 84.50±2.01 & 34.26±0.85 / 27.46±0.86 & 59.99±2.42 / 60.27±2.05 & 71.01±2.11 / 71.36±1.14 \\
Finnish-ModernBERT-large-edu & 401.26 & 70.82±3.40 / 85.02±1.80 & 38.94±2.38 / 32.36±2.23 & 58.85±2.17 / 59.45±1.87 & 70.12±3.29 / 68.40±3.87 \\
Finnish-ModernBERT-large-nce & 401.26 & 66.97±2.66 / 83.38±1.36 & 38.83±2.12 / 32.53±2.09 & 59.65±1.64 / 59.91±1.33 & 70.18±3.77 / 69.85±4.05 \\
Finnish-ModernBERT-large-short & 475.45 & \textbf{75.02±1.35 / 87.29±0.67} & 37.49±2.13 / 30.99±2.20 & 61.67±2.85 / 61.05±2.76 & 71.39±2.10 / 69.44±3.05 \\
Finnish-ModernBERT-large-short-cpt & 475.45 & 73.14±2.62 / 86.27±1.32 & 39.91±1.98 / 33.71±2.00 & 66.01±2.15 / 65.26±1.79 & 72.37±1.45 / 70.93±1.92 \\
Finnish-ModernBERT-large-short-edu & 475.45 & 64.81±5.27 / 82.17±2.59 & 38.67±2.32 / 32.42±2.37 & 65.94±2.17 / 65.66±2.02 & 71.86±1.47 / 70.53±2.16 \\
Finnish-ModernBERT-tiny & 51.65 & 11.31±3.88 / 54.81±2.30 & 27.19±0.82 / 19.54±0.97 & 48.06±2.18 / 49.55±1.87 & 63.73±1.75 / 63.98±1.64 \\
Finnish-ModernBERT-tiny-edu & 51.65 & 8.37±3.26 / 52.59±2.28 & 27.89±0.67 / 20.11±0.82 & 41.71±1.90 / 42.69±1.78 & 63.72±1.75 / 64.38±1.40 \\
Finnish-ModernBERT-tiny-short & 129.11 & 24.95±4.60 / 62.02±2.50 & 30.81±0.71 / 24.17±0.69 & 53.74±2.13 / 55.21±2.11 & 67.57±1.36 / 65.79±2.13 \\
Finnish-ModernBERT-tiny-short-cpt & 129.11 & 24.10±5.74 / 61.88±2.89 & 27.98±1.04 / 21.04±1.09 & 53.90±1.63 / 55.20±1.74 & 67.44±0.83 / 65.56±2.03 \\
Finnish-ModernBERT-tiny-short-edu & 129.11 & 24.36±5.39 / 61.97±2.74 & 28.05±0.77 / 21.10±0.92 & 53.89±2.29 / 55.44±2.12 & 67.34±1.39 / 65.80±2.13 \\
mmBERT-base & 307.53 & 63.62±3.30 / 81.52±1.77 & 44.09±1.64 / 37.17±1.74 & 70.41±1.81 / 70.03±1.63 & 66.91±2.76 / 65.28±3.48 \\
XLM-RoBERTa-large & 561.19 & 72.61±2.84 / 85.79±1.42 & \textbf{47.91±1.23 / 41.40±1.00} & \textbf{79.12±1.13 / 76.69±1.14} & \textbf{75.34±0.60 / 70.16±2.52} \\
\bottomrule
\end{tabular}
\caption{Results for tasks included in EuroEval in the Swedish language. In the metrics header, "MCC" is an abbreviation for Matthews Correlation Coefficient, "Ma" represents macro, "Mi" represents micro, and "EM" represents exact match. The best average scores for monolingual and multilingual models are highlighted in bold.}
\label{tab:euroeval-sv}
\end{table*}

\begin{table*}
\centering
\scriptsize
\setlength{\tabcolsep}{3pt}
\begin{tabular}{l l c c c c}
\textbf{Model} & \textbf{N params (M)} & \textbf{conll-en} & \textbf{scala-en} & \textbf{squad} & \textbf{sst5} \\
\cmidrule(lr){3-6} & & \textit{Mi F1 no misc / Mi F1} & \textit{MCC / Ma F1} & \textit{F1 / EM} & \textit{MCC / Ma F1} \\
\midrule
\multicolumn{6}{l}{\textbf{Monolingual}} \\
DeBERTaV3-base & 184.42 & 91.05±0.53 / 90.46±0.54 & 64.68±1.29 / 81.85±0.67 & 75.68±0.86 / 62.80±0.98 & 62.03±1.05 / 60.52±3.55 \\
DeBERTaV3-large & 434.02 & \textbf{91.46±0.76 / 91.15±0.74} & \textbf{74.30±0.69 / 87.02±0.31} & \textbf{81.39±0.75 / 69.12±0.70} & \textbf{62.65±1.75 / 62.74±2.25} \\
\addlinespace
\midrule
\multicolumn{6}{l}{\textbf{Multilingual}} \\
Finnish-ModernBERT-base & 143.40 & 70.64±2.52 / 72.96±1.99 & 14.04±3.08 / 56.21±1.86 & 29.36±6.50 / 18.20±5.63 & 33.81±3.80 / 46.50±2.77 \\
Finnish-ModernBERT-base-edu & 143.40 & 72.03±1.69 / 74.17±1.30 & 18.63±2.59 / 58.42±1.16 & 24.92±8.63 / 14.88±7.09 & 32.18±2.89 / 47.48±2.25 \\
Finnish-ModernBERT-base-short & 209.35 & 75.95±1.14 / 77.60±0.90 & 28.54±6.94 / 63.14±3.56 & 39.90±7.49 / 26.83±6.13 & 35.03±5.63 / 49.31±4.84 \\
Finnish-ModernBERT-base-short-cpt & 209.35 & 76.00±1.42 / 77.82±1.19 & 19.05±4.51 / 59.15±2.41 & 40.49±8.21 / 27.56±6.91 & 34.78±5.40 / 48.80±4.83 \\
Finnish-ModernBERT-base-short-edu & 209.35 & 74.45±1.05 / 76.34±0.86 & 19.62±5.14 / 58.72±3.08 & 41.91±5.93 / 28.50±5.23 & 40.25±1.58 / 52.06±1.59 \\
Finnish-ModernBERT-large & 401.26 & 79.73±1.29 / 80.90±1.11 & 50.98±3.90 / 74.94±2.06 & 55.98±2.65 / 40.35±2.57 & 37.08±5.53 / 49.38±4.69 \\
Finnish-ModernBERT-large-edu & 401.26 & 79.56±1.76 / 80.98±1.45 & 48.17±3.85 / 73.70±2.05 & 54.77±1.57 / 39.10±1.94 & 35.14±4.42 / 47.51±4.11 \\
Finnish-ModernBERT-large-nce & 401.26 & 79.15±0.60 / 80.20±0.47 & 46.82±5.34 / 72.62±2.64 & 58.70±1.98 / 42.86±1.95 & 38.60±3.48 / 51.67±3.58 \\
Finnish-ModernBERT-large-short & 475.45 & 82.00±1.00 / 82.61±0.89 & \textbf{55.71±3.69 / 77.37±2.02} & 57.98±2.49 / 42.68±2.92 & 44.08±4.02 / 54.98±2.44 \\
Finnish-ModernBERT-large-short-cpt & 475.45 & 80.58±1.37 / 81.38±1.14 & 51.47±7.85 / 75.52±4.01 & 57.51±3.24 / 42.54±3.54 & 47.15±3.84 / 57.53±2.66 \\
Finnish-ModernBERT-large-short-edu & 475.45 & 78.06±0.76 / 78.89±0.65 & 54.38±3.88 / 76.82±2.20 & 59.28±2.46 / 44.09±2.48 & 45.57±1.72 / 54.43±2.04 \\
Finnish-ModernBERT-tiny & 51.65 & 68.71±1.09 / 71.02±0.89 & 4.72±2.12 / 51.47±1.40 & 12.00±0.47 / 4.96±0.43 & 21.24±4.35 / 40.46±2.94 \\
Finnish-ModernBERT-tiny-edu & 51.65 & 67.73±1.66 / 70.23±1.42 & 3.96±1.69 / 51.44±1.08 & 10.99±0.31 / 3.87±0.32 & 21.97±3.27 / 42.10±2.11 \\
Finnish-ModernBERT-tiny-short & 129.11 & 73.77±1.03 / 74.93±0.98 & 16.66±4.59 / 57.55±2.33 & 12.04±0.61 / 4.87±0.51 & 31.64±2.62 / 46.38±2.36 \\
Finnish-ModernBERT-tiny-short-cpt & 129.11 & 73.78±0.62 / 74.93±0.58 & 13.06±1.82 / 56.26±0.93 & 12.04±0.65 / 4.81±0.62 & 30.83±2.93 / 46.58±2.61 \\
Finnish-ModernBERT-tiny-short-edu & 129.11 & 73.53±0.92 / 74.94±0.85 & 12.80±2.58 / 55.93±1.74 & 12.18±0.49 / 4.96±0.47 & 32.02±2.85 / 46.63±2.60 \\
mmBERT-base & 307.54 & 85.43±1.06 / 85.51±1.03 & 47.28±8.78 / 73.15±4.44 & 64.01±2.45 / 48.08±2.54 & 42.97±3.91 / 55.52±3.00 \\
XLM-RoBERTa-large & 561.19 & \textbf{88.74±1.06 / 88.12±0.94} & 34.33±15.56 / 64.04±9.79 & \textbf{70.42±0.84 / 57.34±0.82} & \textbf{58.86±1.33 / 58.07±2.23} \\
\bottomrule
\end{tabular}
\caption{Results for tasks included in EuroEval in the English language. In the metrics header, "MCC" is an abbreviation for Matthews Correlation Coefficient, "Ma" represents macro, "Mi" represents micro, and "EM" represents exact match. The best average scores for monolingual and multilingual models are highlighted in bold.}
\label{tab:euroeval-en}
\end{table*}

The Finnish ModernBERTs performed competitively with other multilingual models on the short context NLU tasks, where XLM-R was the strongest model in most of them. Our models seemed to perform disadvantageously in tasks where token-level predictions are important, such as question answering. This may be due to the large rotary base, which levels out the angle functions, causing less differentiation between close positions.

\subsection{Long context and code evaluation}
 We evaluated the models in single-vector out-of-domain retrieval using the same out-of-domain evaluation procedure as \citet{modernbert}. In contrast to their evaluations, we evaluated the code tasks in an out-of-domain setting. We trained every model on query-positive-negative triplets using a version\footnote{\url{https://huggingface.co/datasets/sentence-transformers/msmarco-co-condenser-margin-mse-sym-mnrl-mean-v1}} of MS-MARCO \cite{msmarco} containing the hard negatives \cite{sbert}. We used 1.25M samples from the data, employing the specified hyperparameters. We did a hyperparameter search for learning rates, $learning\ rate \in \{1e-5, 2e-5, 3e-5, 5e-5, 8e-5,1e-4\}$, selecting the best models based on the mean nDCG@10 metric in four tasks in BEIR \cite{beir}: SciFact \cite{scifact}, FiQA2018 \cite{fiqa}, NFCorpus \cite{nfcorpus}, and TREC-COVID \cite{treccovid}. The results in the short context retrieval on these tasks are provided in Appendix \ref{appendix:d}. After selecting the models, we evaluated them on the long context document retrieval task, MLDR \cite{mldr}, and code tasks CodeSearchNet \cite{hcodesearchnet} and StackOverflowQA \cite{stackoverflowqa} without any further fine-tuning. All retrieval evaluations were run using MTEB \cite{mmteb}. The results in MLDR are presented in Table \ref{tab:mldr-performance} and the results for the code tasks in Table \ref{tab:code-performance}.

\begin{table}[ht]
\small
\centering
\begin{tabular}{lcc}
\toprule
\textbf{Model} & \textbf{MLDR} \\
\midrule
\multicolumn{2}{l}{\textbf{Monolingual Models}} \\
\midrule
GTE-en-MLM-large* & \textbf{0.364} \\
ModernBERT-large* & 0.343 \\
RoBERTa-base* & 0.226 \\
DeBERTaV3-large* & 0.071 \\
\midrule
\multicolumn{2}{l}{\textbf{Multilingual Models}} \\
\midrule
Finnish-ModernBERT-base & 0.218 \\
Finnish-ModernBERT-base-edu & 0.255 \\
Finnish-ModernBERT-base-short & 0.259 \\
Finnish-ModernBERT-base-short-cpt & 0.259 \\
Finnish-ModernBERT-base-short-edu & 0.208 \\
Finnish-ModernBERT-large & 0.302 \\
Finnish-ModernBERT-large-edu & 0.306 \\
Finnish-ModernBERT-large-short & \textbf{0.311} \\
Finnish-ModernBERT-large-short-cpt & 0.287 \\
Finnish-ModernBERT-large-short-edu & 0.302 \\
Finnish-ModernBERT-tiny & 0.151 \\
Finnish-ModernBERT-tiny-edu & 0.135 \\
Finnish-ModernBERT-tiny-short & 0.196 \\
Finnish-ModernBERT-tiny-short-cpt & 0.153 \\
Finnish-ModernBERT-tiny-short-edu & 0.197 \\
mmBERT-base & 0.072 \\
XLM-RoBERTa-large & 0.166 \\
\bottomrule
\end{tabular}
\caption{Model performance in out-of-domain retrieval MLDR using the nDCG@10 metric. Results with "*" are reported by \protect{\citet{modernbert}}. The best results for each task and the best average are highlighted in bold for both monolingual and multilingual models.}
\label{tab:mldr-performance}
\end{table}

\begin{table*}
\centering
\begin{tabular}{lccc}
\toprule
\textbf{Model} & \textbf{CSN} & \textbf{SQA} & \textbf{Average} \\
\midrule
\multicolumn{4}{l}{\textbf{Monolingual Models In-Domain performance}} \\
\midrule
\textit{GTE-en-MLM-base} & 0.449 & 0.714 & 0.582 \\
\textit{ModernBERT-large} & \textbf{0.595} & \textbf{0.839} & \textbf{0.717} \\
\midrule
\multicolumn{4}{l}{\textbf{Multilingual Models Out-of-Domain performance}} \\
\midrule
Finnish-ModernBERT-base & 0.361 & 0.536 & 0.448 \\
Finnish-ModernBERT-base-edu & 0.375 & 0.515 & 0.445 \\
Finnish-ModernBERT-base-short & 0.377 & 0.547 & 0.462 \\
Finnish-ModernBERT-base-short-cpt & 0.366 & 0.550 & 0.458 \\
Finnish-ModernBERT-base-short-edu & 0.349 & 0.538 & 0.443 \\
Finnish-ModernBERT-large & 0.461 & 0.608 & 0.535 \\
Finnish-ModernBERT-large-edu & 0.447 & 0.612 & 0.529 \\
Finnish-ModernBERT-large-short & \textbf{0.467} & \textbf{0.635} & \textbf{0.551} \\
Finnish-ModernBERT-large-short-cpt & 0.455 & 0.619 & 0.537 \\
Finnish-ModernBERT-large-short-edu & 0.452 & 0.621 & 0.537 \\
Finnish-ModernBERT-tiny & 0.219 & 0.491 & 0.355 \\
Finnish-ModernBERT-tiny-edu & 0.217 & 0.488 & 0.353 \\
Finnish-ModernBERT-tiny-short & 0.333 & 0.549 & 0.441 \\
Finnish-ModernBERT-tiny-short-cpt & 0.316 & 0.534 & 0.425 \\
Finnish-ModernBERT-tiny-short-edu & 0.323 & 0.548 & 0.436 \\
mmBERT-base & 0.454 & 0.593 & 0.523 \\
XLM-RoBERTa-large & 0.395 & 0.562 & 0.478 \\
\bottomrule
\end{tabular}
\caption{Model performance across different code tasks using the nDCG@10 metric. The best results for each task and the best averages are highlighted in bold for both monolingual and multilingual models. In-domain results, highlighted in italics, are reported by \protect{\citet{modernbert}}.}
\label{tab:code-performance}
\end{table*}

The evaluation trend was that models trained on the edu data during the annealing phase performed a bit worse on both out-of-domain retrieval tasks. This may simply reflect the fact that the baseline annealing data contained more English overall.

\section{Discussion and conclusions}

We reported on pretraining the Finnish ModernBERTs for languages relevant in Finland. We explored the use of edu data in the annealing phase and evaluated our models extensively. The evaluation results show that the Finnish ModernBERTs are competitive or superior to existing models.

Finnish ModernBERTs were the strongest multilingual encoder models in out-of-domain retrieval tasks. The out-of-domain retrieval performance of the other multilingual models in MLDR was more than 14 points lower than that of the leading Finnish ModernBERT model. The 7th percentile of the MLDR documents' lengths was 1125 tokens, the 50th percentile was 2269 tokens, and the 93rd percentile was 7720 when using a tokenizer with a fertility close to 1 in English, indicating that sequence length extrapolation and performance near the maximum sequence length remain unknown. However, we conclude that the context extension was successful, as the models can retrieve beyond the stable phase sequence length of 1024.\footnote{As the 7th percentile contains 14 documents, retrieving only them in perfect rank would yield $ nDCG@10=0.07$. If a system were to retrieve the 14 documents perfectly, to obtain a score of $0.307$ system would additionally need to retrieve 125 documents with rank 5}. The usefulness of the non-English edu annealing data for enhancing the model performance in Finnish and Swedish remains unconfirmed, as performance fluctuates across evaluation datasets in short context evaluations.

\section*{Limitations}
We employed a pretraining method that compiles findings from various preceding research. We did not have the resources to run distinct ablation trials to assess the impact of these choices (e.g., language distributions, data oversampling, rotary base). While our results demonstrate that this pretraining method can be used to train capable multilingual models, the research is limited and leaves open questions.

The models were not evaluated on the Latin or Northern S{\'a}mi benchmarks. The inclusion of Latin was not intended to develop better models for Latin itself, but rather to support the models' representation capability in domains that utilize Latin terms. To the best of our knowledge, the only evaluation resource for Northern S{\'a}mi is Tatoeba \cite{tatoeba}, which contains 68 English sentences and Northern S{\'a}mi translations. Evaluating an encoder model in translation is not appropriate, as the model's generation capabilities are limited. 

We evaluated the models extensively across multilingual short context NLU tasks. The models' performance in multilingual long context NLU tasks is unknown, and we encourage practitioners to test these models on long context tasks where optimized representations of Finnish, Swedish, English, and Northern S{\'a}mi are needed.

The training code is a compromise between efficiency and flexibility. While using a custom code helped in inserting experimental options, the training code did not utilize all possible efficiency improvements, such as sequence packing \cite{packing1,packing2} and model parallelism \cite{megatron-lm}.

\section*{Ethical and environmental considerations}
We acknowledge the potential negative consequences of openly distributing models that can be used to create representations meaningful to humans. These consequences include, but are not limited to, an increase in search-based surveillance and cyber violence, such as researching and publishing private information on the internet to expose and shame the target. Since the training data includes sources regarded as biased and harmful, the models might mirror these biased properties. We did not filter the harmful content from the data to serve various use cases. The representations produced by the models should not be used without caution and without evaluating their effects on vulnerable population groups. These limitations are communicated openly at the distribution site.

Training large transformer-based models requires immense computational resources, resulting in significant energy consumption, carbon emissions, and costs. The growing awareness of these impacts has prompted increasing attention to the energy and resource demands of model development and deployment \cite{bakhtiarifard2025climate}.
An eco-efficient model design focuses on minimizing energy, water, and resource consumption throughout the model lifecycle. Prior work has shown that using more energy-efficient hardware, optimizing data center location, and leveraging renewable energy sources can significantly reduce emissions \cite{nn-carbon}.

The computational costs of larger models, combined with the environmental impact, advocate for the use of specialized smaller models when possible, instead of relying on in-context learning with much larger generative GPT-style decoder-only models \cite{lehman2023we,chen2024evaluating}.

Eco-efficient design, proactive e-waste management, and transparent system reporting are central approaches to sustainable model development and deployment. Our ModernBERT models incorporated several eco-efficient design principles that may reduce their environmental impact. The encoder-only architecture inherently requires less compute than decoder models of comparable performance, while our range of model sizes (51M-475M parameters) enables the deployment of right-sized models for specific tasks, avoiding unnecessary computational overhead.  Additionally, training in Finland leverages one of Europe's cleanest energy grids with a high renewable content compared to average data center locations \cite{energy-finland}.

\section*{Author Contributions}

We report author contributions following CRediT taxonomy\footnote{\url{https://credit.niso.org/}}. \textbf{Akseli Reunamo}: Conceptualization, Data Curation, Formal Analysis, Funding Acquisition, Investigation, Methodology, Project Administration, Software, Validation, Visualization, Writing - Original Draft Preparation, Writing - Review \& Editing. \textbf{Hans Moen}: Conceptualization, Supervision, Writing - Original Draft Preparation, Writing - Review \& Editing. \textbf{Laura-Maria Peltonen}: Conceptualization, Supervision, Writing - Original Draft Preparation, Writing - Review \& Editing. \textbf{Sampo Pyysalo}: Conceptualization, Data Curation, Funding Acquisition, Resources, Supervision, Validation, Writing – Review \& Editing.

\section*{Acknowledgments}
We acknowledge CSC, IT Center for Science, Finland, for awarding this project access to the LUMI supercomputer, owned by the EuroHPC Joint Undertaking, hosted by CSC (Finland) and the LUMI consortium. We acknowledge the HPLT-project for supporting this research. This project has received funding from the European Union’s Horizon Europe research and innovation programme under grant agreement No. 101070350, and it has also received funding from the Finnish Cultural Foundation. 
\bibliography{custom}

\appendix
\onecolumn
\section{Training details}
\label{appendix:a}
All models were trained using the same schedulers during the stable, context extension, and annealing phases (Table \ref{tab:training_schedules}). During the context extension phase, the sequence length was increased to the final length in six steps, where each length was trained for the same number of steps. We adopted this strategy from Llama 3 training \cite{llama31}. Similarly, this kind of expansive method was used in DeepSeek Coder \cite{deepseekcoder}. In the context extension and annealing phases, the data were always randomly shuffled when training was continued from a checkpoint or a new sequence length was introduced. We assumed that, since datasets were already oversampled, a few additional repeats would not negatively impact performance. Training hyperparameters for each model size are provided in Table \ref{tab:training_hyp}. 

\begin{table}[ht]
    \centering
    \begin{tabular}{llllll}
        & \textbf{Scheduler} & \textbf{Start Step} & \textbf{Steps} & \textbf{\%} \\ \hline
        \multirow{4}{*}{\textbf{Stable}} 
        & LR warmup & 0 & 1380 & 1 \\
        & Batch size warmup & 0 & 4002 & 2.9  \\ 
        & LR constant & 1380 & 115920 & 84 \\
        & Global layer RoPE rotary base $10000$ & 0 & 117300 & 85 \\ \hline
        
        \multirow{3}{*}{\textbf{Context Extension}} 
        & LR second constant & 117300 & 16560 & 12 \\
        & Context extension data & 117300 & 16560 & 12 \\
        & Global layer RoPE rotary base $1000000$ & 117300 & 20700 & 15 \\ \hline
        
        \multirow{2}{*}{\textbf{Annealing}} 
        & LR Decay & 133860 & 4140 & 3 \\
        & Annealing data & 133860 & 4140 & 3  \\ \hline
    \end{tabular}
    \caption{Number of steps for each scheduler in different training phases. }
    \label{tab:training_schedules}
\end{table}

\begin{table}[ht]
    \centering
    \begin{tabular}{llllll}
    \textbf{Model size} & \textbf{S LR} & \textbf{C LR} & \textbf{Weight decay} & \textbf{AdamW $\beta$} & \textbf{AdamW $\epsilon$} \\
    \hline
    tiny                &   $8e-4$    & $5e-4$              & $1e-5$                 & $(0.9,0.98)$ &  $1e-6$ \\  
    base                &   $5e-4$    & $3e-4$              & $1e-5$                 & $(0.9,0.98)$ &  $1e-6$ \\                
    large               &   $3e-4$    & $5e-5$              & $1e-6$                 & $(0.9,0.98)$ &  $1e-6$ \\ 
    \hline
    \end{tabular}
    \caption{Training hyperparameters. "S" refers to the Stable phase and "C" to the Context extension phase.}
    \label{tab:training_hyp}
\end{table}

The number of tokens used to train the models is provided in Table \ref{tab:tokens}. The models trained with optimal vocabulary and annealed with baseline annealing data received twice as many tokens during the annealing phase as those trained with the edu annealing data.

\begin{table}[ht]
    \small
    \centering
    \begin{tabular}{lllll|l|llll}
    \textbf{Model} & \textbf{W} & \textbf{S} & \textbf{C} &  \textbf{A} & \textbf{Total} & \textbf{W Avg Bs} & \textbf{S Avg Bs} &  \textbf{C Avg Bs} & \textbf{A Avg Bs}   \\
    \hline
    tiny                & 6.5B & 370.0B & 53.4B & 18.0B & 447.9B & 1.6M  & 3.3M  & 3.2M  & 4.4M \\
    tiny-edu            & 6.5B & 370.0B & 53.4B & 9.4B  & 439.3B & 1.6M  & 3.3M  & 3.2M & 2.3M  \\  
    tiny-short          & 5.6B & 320.4B & 31.3B & 4.9B  & 362.2B & 1.4M  & 2.8M & 1.9M  & 1.2M \\  
    tiny-short-edu      & 5.6B & 320.4B & 31.3B & 4.9B  & 362.2B & 1.4M  & 2.8M & 1.9M & 1.2M \\  
    base                & 6.2B & 353.3B & 39.2B & 11.4B & 410.2B & 1.6M  & 3.1M & 2.4M & 2.8M \\
    base-edu            & 6.2B & 353.3B & 39.2B & 5.9B  & 404.7B & 1.6M  & 3.1M & 2.4M & 1.4M \\ 
    base-short          & 5.7B & 320.4B & 34.1B & 4.9B  & 365.0B & 1.4M  & 2.8M  & 2.1M  & 1.2M \\                
    base-short-edu      & 5.7B & 320.4B & 34.1B & 4.9B  & 365.0B  & 1.4M  & 2.8M  & 2.1M  & 1.2M \\          
    large               & 6.1B & 343.9B & 38.1B & 11.1B  & 399.2B  & 1.5M  & 3.0M & 2.3M & 2.7M \\
    large-edu           & 6.1B & 343.9B & 38.1B & 5.6B  & 393.8B & 1.5M  & 3.0M  & 2.3M & 1.4M \\
    large-short         & 5.7B & 320.3B & 31.3B & 4.9B  & 362.2B & 1.4M & 2.8M  & 1.9M  & 1.2M \\
    large-short-edu     & 5.7B & 320.3B & 31.3B & 4.9B  & 362.2B & 1.4M & 2.8M  & 1.9M  & 1.2M \\
    \hline
    \end{tabular}
    \caption{Training tokens for each Finnish ModernBERT model. "W" refers to the warm-up phase, "S" refers to the Stable phase, "C" to the Context extension phase, and "A" to the annealing phase. "Avg" refers to the average, and "Bs" to the batch size in tokens. The suffixes "B" and "M" represent billions and millions.}
    \label{tab:tokens}
\end{table}

\section{Datasets used in pretraining}
\label{appendix:b}
All datasets (Table \ref{tab:datasets}) were converted into JSONL format, and train splits were used when relevant; if not explicitly reported, no alternative splits were used. Some datasets contain additional processing steps, which are documented either in the "Notes" section or in the relevant literature. Wikipedia dumps were processed using Wikiectrator \cite{Wikiextractor2015}. For cross-lingual data, each entry was prefixed with a translation instruction (Table \ref{tab:translation}). 
\begin{table}[ht]
    \centering
    \begin{tabular}{ll}
        \textbf{Languages}  & \textbf{Instruction}\\ \hline
        eng-fin  & Translate into Finnish:  \\ 
        eng-sme  & Translate into Northern S{\'a}mi: \\ 
        eng-swe  & Translate into Swedish:  \\
        fin-eng  & Käännä englanniksi:  \\ 
        fin-sme  & Käännä pohjoissaameksi: \\ 
        fin-swe  & Käännä ruotsiksi:  \\
        sme-fin  & Jorgal suomagillii:  \\ 
        sme-eng  & Jorgal e\textipa{N}gelasgillii: \\ 
        sme-swe  & Jorgal ruo\tstroke agillii:  \\
        swe-fin  & {\"O}versätt till finska:  \\ 
        swe-eng  & {\"O}versätt till engelska:  \\
        swe-sme  & {\"O}versätt till nordsamiska: \\
    \hline
    \end{tabular}
    \caption{Translation instruction for each language pair. Language codes are ISO 639-3 codes.}
    \label{tab:translation}
\end{table}

\begin{sidewaystable}[ht]
\tiny      
\begin{tabular}{llllllrlrllrl}
\toprule
Lang    &                Dataset &                           Notes & References &     I &     D &  R S-D\% & PII-removed &  R D-P\% &     F &     P &     S & Final \\
\midrule
    code &             Starcoder &     GitHub issues               &          \citenum{starcoder} & 15.5B & 15.4B &   -0.54 &       15.4B &  -0.014 &       & 15.4B &  0.83 &        12.8B \\
    code &                SmolLM & PythonEdu score 5               &          \citenum{smollmcorpus} & 45.5M & 45.5M &    0.00 &       45.5M &  -0.011 &       & 45.5M & 30.00 &         1.4B \\
     eng &      Brithish Library & eng subset, $std\_wc\_ocr<0.14$ &          \citenum{britishlibrary} &  - &       &       - &             &       - &   1.9B    &  1.9B &  1.00 &         1.9B \\
     eng &              Europarl & eng subset, parsed January 2024 & \citenum{thepile,pileeuroparl} & 62.6M &       &       - &             &       - &       & 62.6M &  5.00 &         0.3B \\
     eng & FineWeb-Edu fortified &                 -               &          \citenum{fineweb-edu-fortified,fineweb} & 69.6B &       &       - &       69.5B &       - &       & 69.5B &  0.50 &        34.8B \\
     eng &  Natural Instructions &              v2.8               &       \citenum{naturalinstructions,naturalinsprocessed} &  0.7B &       &       - &             &       - &       &  0.7B &  1.00 &         0.7B \\
     eng &                 pes2o &                 -               &          \citenum{peS2o} & 51.9B &       &       - &             &       - &       & 51.9B &  0.13 &         6.8B \\
     eng &        PubMed Central &                 -               &          \citenum{thepile,pile-pubmedcentral} & 22.1B &       &       - &             &       - &       & 22.1B &  0.10 &         2.2B \\
     eng &      PubMed Abstracts &                 -               &          \citenum{thepile,pilepubmedabstract} &  3.8B &       &       - &             &       - &       &  3.8B &  1.00 &         3.8B \\
     eng &             Wikipedia & Dump 20241101 eng               &          \citenum{wikipedia} &  3.4B &       &       - &             &       - &       &  3.4B &  9.00 &        30.3B \\
     fin &                 CC-fi &                 -               &  \citenum{fingpt}    & 10.9B & 10.8B &   -0.84 &       10.8B &   0.011 &       & 10.8B &  4.00 &        43.4B \\
     fin &              CulturaX &        fin subset               & \citenum{culturax}            & 17.0B & 16.9B &   -0.12 &       16.9B &  -0.006 &       & 16.9B &  3.70 &        62.7B \\
     fin &              HPLT 2.0 &        fin, cleaned subset      & \citenum{hpltv2}              & 23.3B & 19.2B &  -17.53 &       19.2B &  -0.029 &       & 19.2B &  3.70 &        71.0B \\
     fin &                 NLFCL &        fin subset               & \citenum{nlfcl-fi}           & 21.6M &       &       - &             &       - & 21.4M & 21.4M &  6.00 &         0.1B \\
     fin &              Europarl &        fin subset & \citenum{thepile,pileeuroparl}  & 41.6M &       &       - &             &       - &       & 41.6M &  6.00 &         0.2B \\
     fin &               Lönnrot &            FinGPT & \citenum{fingpt,lonnrot} &  0.1B &       &       - &             &       - &       &  0.1B &  6.00 &         0.8B \\
     fin &             Reddit-Fi &            FinGPT & \citenum{fingpt} &  0.1B &       &       - &             &       - &       &  0.1B &  6.00 &         0.7B \\
     fin &               Suomi24 &            FinGPT & \citenum{fingpt,suomi24-2001-2020}&  3.3B &       &       - &        3.3B &       - &       &  3.3B &  6.00 &        19.6B \\
     fin &             Wikipedia & Dump 20241101 fin & \citenum{wikipedia} &  0.1B &       &       - &             &       - &       &  0.1B & 30.00 &         3.9B \\
     fin &              Yle news &            FinGPT & \citenum{fingpt,ylenews-fi-2011-2018,ylenews-fi-2011-2018-selko,ylenews-fi-2019-2020,ylenews-fi-2019-2020-selko} &  0.2B &       &       - &             &       - &       &  0.2B & 30.00 &         6.7B \\
     fin &              Ylilauta &                 - & \citenum{ylilauta} & 28.1M & 17.6M &  -37.31 &       17.6M &   0.021 &       & 17.6M &  5.00 &        88.1M \\
     lat &              CulturaX &        lat subset & \citenum{culturax} & 31.5M &       &       - &       31.5M &       - &       & 31.5M & 30.00 &         0.9B \\
     sme &               Glot500 &        sme subset & \citenum{glot500} &  3.5M &  3.5M &    0.00 &        3.5M &  -0.015 &       &  3.5M & 30.00 &         0.1B \\
     sme &             saami-web &                 - & \citenum{saami-web} & 16.7M & 16.7M &   -0.08 &       16.7M &  -0.055 &       & 16.7M & 30.00 &         0.5B \\
     sme &                  SALT &                 - & \citenum{salt-data,salt-article} & 17.7M & 15.3M &  -13.52 &       15.3M &  -0.031 &       & 15.3M & 30.00 &         0.5B \\
     swe &              CulturaX &        swe subset & \citenum{culturax} & 28.7B &       &       - &       28.7B &       - &       & 28.7B &  1.09 &        31.3B \\
     swe &              Europarl &        swe subset, parsed January 2024 & \citenum{thepile,pileeuroparl} & 52.4M &       &       - &             &       - &       & 52.4M &  5.00 &         0.3B \\
     swe &                  FSTC &        spoken part of corpus excluded & \citenum{fstc-source} &  2.3M &       &       - &             &       - &       &  2.3M &  5.00 &        11.3M \\
     swe &              HPLT 2.0 &        swe, cleaned subset & \citenum{hpltv2} & 46.8B & 35.8B &  -23.62 &       35.8B &  -0.045 &       & 35.8B &  1.05 &        37.5B \\
     swe &                 NLFCL &        swe subset & \citenum{nlfcl-sv} & 14.0M &       &       - &             &       - &       & 14.0M &  5.00 &        70.1M \\
     swe &             Wikipedia & Dump 20241101 swe & \citenum{wikipedia,Wikiextractor2015}         - &  0.3B &       &       - &             &       - &       &  0.3B & 30.00 &         8.1B \\
     swe &              Yle news &        swe subset & \citenum{ylenews-sv-2012-2018,ylenews-sv-2012-2018}& 95.1M & 95.1M &   -0.04 &             &       - &       & 95.1M & 30.00 &         2.9B \\
   xling &               Tatoeba &           eng-fin v2023-09-26 & \citenum{tatoeba} &  1.1B &       &       - &             &       - &       &  1.1B &  0.62 &         0.7B \\
   xling &                  OPUS &           eng-sme, wikimedia v20230407 & \citenum{opus} &  5.0K &       &       - &             &       - &       &  5.0K & 30.00 &       150.4K \\
   xling &               Tatoeba &           eng-swe v2023-09-26 & \citenum{tatoeba} &  1.2B &       &       - &             &       - &       &  1.2B &  0.57 &         0.7B \\
   xling &               Tatoeba &           fin-eng v2023-09-26 &  \citenum{tatoeba} &  1.1B &       &       - &             &       - &       &  1.1B &  0.62 &         0.7B \\
   xling &                  OPUS &           fin-sme, wikimedia v20230407 & \citenum{opus}  & 12.8K &       &       - &             &       - &       & 12.8K & 30.00 &       382.7K \\
   xling &               Tatoeba &           fin-swe v2023-09-26 &          \citenum{tatoeba} &  0.1B &       &       - &             &       - &       &  0.1B &  5.70 &         0.7B \\
   xling &                  OPUS &           sme-eng, wikimedia v20230407 & \citenum{opus} &  4.8K &       &       - &             &       - &       &  4.8K & 30.00 &       145.0K \\
   xling &                  OPUS &           sme-fin, wikimedia v20230407 & \citenum{opus} & 12.8K &       &       - &             &       - &       & 12.8K & 30.00 &       382.7K \\
   xling &                  OPUS &           sme-swe, wikimedia v20230407 & \citenum{opus} &  0.9K &       &       - &             &       - &       &  0.9K & 30.00 &        25.6K \\
   xling &               Tatoeba &           swe-eng v2023-09-26 & \citenum{tatoeba} &  1.2B &       &       - &             &       - &       &  1.2B &  0.58 &         0.7B \\
   xling &               Tatoeba &           swe-fin v2023-09-26 & \citenum{tatoeba} &  0.1B &       &       - &             &       - &       &  0.1B &  5.70 &         0.7B \\
   xling & OPUS & swe-sme, wikimedia v20230407 & \citenum{opus} &  0.9K &       &       - &             &       - &       &  0.9K & 30.00 &        26.3K \\
\bottomrule
\end{tabular}
\caption{Datasets and their sizes in tokens after each step of processing. In the "Lang" column, "code" refers to programming languages and "xling" to cross-lingual data; other codes are ISO 639-3 codes. In the "Dataset" column, "NLFCL" is an abbreviation for Classics Library of the National Library of Finland, and "FSTC" refers to The Finland-Swedish Text Corpus. The suffixes "B", "M", and "K" in numerical columns refer to billions, millions, and thousands, respectively. "-" implies that this operation was not done or is not relevant for a dataset. Column "I" refers to initial token count, column "D" refers to tokens after deduplication, column "R I-D\%" refers to reduction between initial and deduplication, column "R D-P\%" referes to reduction between deduplicated and PII-removed, column "F" refers to tokens tokens after heuristic filtering, column "P" refers to token count after all processing, and column "S" to sampling factor.}
\label{tab:datasets}
\end{sidewaystable}

\section{Context extension and annealing data}
\label{appendix:c}
The sequence length distribution of the context extension data is provided in Table \ref{tab:c_length_dist}. Sources and token quantities of context extension data are provided in Table \ref{tab:c_dist}. Sources and token quantities of the baseline annealing data are described in Table \ref{tab:b_dist}. Sources and token quantities of the edu annealing data are described in Table \ref{tab:edu_dist}. For dataset descriptions, refer to Appendix \ref{appendix:b}.

\begin{table}[ht]
    \centering
    \begin{tabular}{ll}
        \textbf{Sequence length}  & \textbf{\%}\\ \hline
        <1K  & 21.01  \\ 
        1K-10K  & 77.56\\ 
        10K-16K  & 1.03  \\
        >16K & 0.4 \\
    \hline
    \end{tabular}
    \caption{Context length extension data document length distribution. "K" refers to thousands.}
    \label{tab:c_length_dist}
\end{table}

\begin{table}[ht]
    \centering
    \begin{tabular}{lll|l}
        \textbf{Lang}  & \textbf{Source} & \textbf{Tokens} & \textbf{\%}\\
        \hline
        code & SmolLM Python-edu & 998.3M & 0.29 \\
        code & Starcoder Github Issues & 8.6B & 2.52 \\
        \hline
        code & \textbf{Total} & 9.6B & 2.81 \\
        \hline
        fin  & CC-fi & 42.5B & 12.42 \\
        fin  & Europarl & 725.4M  & 0.21 \\
        fin  & HPLT 2.0 & 41.3B  & 12.08 \\
        fin  & Lönnrot & 2.2B  & 0.65 \\
        fin  & Reddit-Fi & 58.9M  & 0.02 \\
        fin  & Suomi24 & 1.8B  & 0.52 \\
        fin  & Yle & 2.20B   & 0.65 \\
        fin  & Ylilauta & 9.9M   & 0.00 \\
        \hline
        fin & \textbf{Total} & 90.8B & 26.55 \\
        \hline
        eng  & Fineweb-edu-fortied & 63.8B  & 18.64 \\
        eng  & Brithish Library & 253.5M  & 0.07 \\
        eng  & Natural Instructions & 248.6M  & 0.07 \\
        eng  & pes2o & 17.4B & 5.10 \\ 
        eng  & PubMed Central & 6.60B & 1.93  \\
        eng  & PubMed Abstracts & 317.3M  & 0.09 \\
        eng  & Wikipedia & 46.10B & 13.47 \\
        \hline
        eng & \textbf{Total} & 134.7B & 39.37 \\
        \hline
        lat  & CulturaX & 2.1B & 0.61 \\
        \hline
        lat & \textbf{Total} & 2.1B & 0.61 \\
        \hline
        sme  & Glot500 & 9.6M & 0.00  \\
        sme  & saami-web  & 658.3M & 0.19  \\
        sme  & SALT & 42.2M & 0.01  \\
        \hline
        sme & \textbf{Total} & 710.1M & 0.20 \\
        \hline
        swe &  CulturaX & 42.0B & 12.28 \\
        swe &  Europarl & 764.3M & 0.22 \\
        swe  & HPLT 2.0 & 59.5B  & 17.38 \\
        swe &  Yle & 1.4B  & 0.42 \\
        swe &  NLFCL & 207.0M & 0.06 \\
        \hline
        swe & \textbf{Total} & 103.9B & 30.36 \\
        \hline
        xling &  fin-eng & 54.9M & 0.02 \\
        xling &  fin-swe & 82.2M & 0.02 \\
        xling &  en-sme & 8.5K & 0.02 \\
        xling &  en-swe & 54.8M & 0.02 \\
        xling &  sme-eng & 8.2K & 0.00 \\
        xling &  sme-fin & 31.7K & 0.00 \\
        xling &  sme-swe & 1.3K & 0.00 \\
        xling &  swe-fin & 87.2M & 0.03 \\
        xling &  swe-sme & 1.4K & 0.00 \\
        \hline
        xling & \textbf{Total} & 279.2M & 0.11 \\
        \hline
        \textbf{Total} &  & \textbf{342.11B}  & \textbf{100} \\
    \hline
    \end{tabular}
    \caption{Sources and token counts of context extension data. In the "Lang" column, "code" refers to programming languages and "xling" to cross-lingual data; other codes are ISO 639-3 codes. The suffixes "B", "M", and "K" in the "Tokens" column refer to billions, millions, and thousands, respectively.}
    \label{tab:c_dist}
\end{table}

\begin{table}[ht]
    \centering
    \begin{tabular}{lll|l}
        \textbf{Language}  & \textbf{Source} & \textbf{Tokens} & \textbf{\%}\\
        \hline
        code & SmolLM Python-edu & 998.3M & 0.69 \\
        \hline
        code & \textbf{Total} & 998.3M & 0.69 \\
        \hline
        fin  & Europarl & 725.4M  & 0.50\\
        fin  & Lönnrot & 2.2B  & 1.54 \\
        fin  & Yle & 2.2B   & 1.55 \\
        \hline
        fin & \textbf{Total} & 5.1B & 3.59 \\
        \hline
        eng  & Fineweb-edu-fortied & 63.8B  & 44.39 \\
        eng  & Brithish Library & 253.5M  & 0.18 \\
        eng  & pes2o & 17.4B & 12.14 \\ 
        eng  & PubMed Central & 6.6B & 4.60  \\
        eng  & PubMed Abstracts & 317.3M  & 0.22 \\
        eng  & Wikipedia & 46.1B & 32.06 \\
        \hline
        eng & \textbf{Total} & 134.5B & 93.59 \\
        \hline
        sme  & saami-web  & 658.3M & 0.46  \\
        \hline
        sme & \textbf{Total} & 658.3M & 0.46 \\
        \hline
        swe &  Europarl & 764.3M & 0.53 \\
        swe &  Yle & 1.4B  & 1.00 \\
        swe &  NLFCL & 207.0M & 0.14 \\
        \hline
        sme & \textbf{Total} & 2.4B & 1.67 \\
        \hline
        \textbf{Total} &  & \textbf{143.69}  & \textbf{100} \\
    \hline
    \end{tabular}
    \caption{Sources and token counts of baseline annealing data. The suffixes "B", "M", and "K" in the "Tokens" column refer to billions, millions, and thousands, respectively.}
    \label{tab:b_dist}
\end{table}

\begin{table}[ht]
    \centering
    \begin{tabular}{llll}
        \textbf{Language}  & \textbf{Source} & \textbf{Tokens} (B) & \textbf{\%}\\
        \hline
        code & SmolLM Python-edu & 0.04 & 0.8 \\
        fin  & HPLT 2.0 cleaned &  2.7 & 54.9 \\ 
        eng  & Fineweb-edu-fortied & 0.90 & 18.3 \\ 
        sme  & saami-web  & 0.49 & 10.0 \\
        swe &  HPLT 2.0 cleaned & 0.79 & 16.0 \\
        \hline
        \textbf{Total} &  & \textbf{4.92} & \textbf{100} \\
    \hline
    \end{tabular}
    \caption{Sources and token counts of edu annealing data. "B" refers to billions.}
    \label{tab:edu_dist}
\end{table}

\section{Short context retrieval results}
\label{appendix:d}
The models' performances on the selected set of BEIR tasks are presented in Table \ref{tab:performance-bier}. The best learning rates for the models trained by us are presented in Table \ref{tab:short_context_lrs}

\begin{table}[ht]
    \centering
    \begin{tabular}{cc}
    \textbf{Model}& \textbf{Learning rate} \\
    \hline
    Finnish-ModernBERT-tiny & 0.0001 \\
    Finnish-ModernBERT-tiny-edu & 8e-05 \\
    Finnish-ModernBERT-tiny-short & 8e-05 \\
    Finnish-ModernBERT-tiny-short-edu & 8e-05 \\
    Finnish-ModernBERT-tiny-short-cpt & 5e-05 \\
    Finnish-ModernBERT-base & 8e-05 \\
    Finnish-ModernBERT-base-edu & 0.0001 \\
    Finnish-ModernBERT-base-short & 2e-05 \\
    Finnish-ModernBERT-base-short-edu & 3e-05 \\
    Finnish-ModernBERT-base-short-cpt & 0.0001 \\
    Finnish-ModernBERT-large & 3e-05 \\
    Finnish-ModernBERT-large-edu & 3e-05 \\
    Finnish-ModernBERT-large-short & 5e-05 \\
    Finnish-ModernBERT-large-short-edu & 5e-05 \\
    Finnish-ModernBERT-large-short-cpt & 2e-05 \\
    mmBERT-base & 1e-05 \\
    XLM-RoBERTa-large & 2e-05 \\
    \hline
    \end{tabular}
    \caption{Best learning rates for the subset of BEIR tasks.}
    \label{tab:short_context_lrs}
\end{table}

\begin{table}[ht]
\centering
\small
\begin{tabular}{lcccc|c}
\toprule
\textbf{Model} & \textbf{FiQA2018} & \textbf{NFCorpus} & \textbf{SciFact} & \textbf{TRECCOVID} & \textbf{Average} \\
\midrule
\multicolumn{6}{l}{\textbf{Monolingual}} \\
\midrule
GTE-en-MLM base* & 0.360 & 0.351 & 0.715 & 0.694 & 0.530 \\
GTE-en-MLM large* & 0.396 & 0.352 & 0.724 & 0.672 & 0.536 \\
ModernBERT base* & 0.380 & 0.352 & 0.730 & 0.805 & 0.567 \\
ModernBERT large* & \textbf{0.403} & \textbf{0.360} & \textbf{0.732} & \textbf{0.813} & \textbf{0.577} \\
\midrule
\multicolumn{6}{l}{\textbf{Multilingual}} \\
\midrule
Finnish-ModernBERT-base & 0.211 & 0.201 & 0.532 & 0.699 & 0.411 \\
Finnish-ModernBERT-base-edu & 0.213 & 0.202 & 0.529 & 0.685 & 0.407 \\
Finnish-ModernBERT-base-short & 0.240 & 0.210 & 0.544 & 0.719 & 0.428 \\
Finnish-ModernBERT-base-short-cpt & 0.250 & 0.216 & 0.529 & 0.720 & 0.429 \\
Finnish-ModernBERT-base-short-edu & 0.237 & 0.216 & 0.541 & 0.716 & 0.428 \\
Finnish-ModernBERT-large & 0.285 & 0.238 & 0.597 & 0.759 & \textbf{0.470} \\
Finnish-ModernBERT-large-edu & 0.276 & 0.238 & 0.593 & 0.736 & 0.461 \\
Finnish-ModernBERT-large-short & 0.284 & 0.226 & 0.602 & 0.738 & 0.463 \\
Finnish-ModernBERT-large-short-cpt & 0.283 & 0.219 & 0.578 & \textbf{0.764} & 0.461 \\
Finnish-ModernBERT-large-short-edu & 0.280 & 0.223 & \textbf{0.615} & 0.752 & 0.468 \\
Finnish-ModernBERT-tiny & 0.182 & 0.189 & 0.517 & 0.590 & 0.369 \\
Finnish-ModernBERT-tiny-edu & 0.186 & 0.186 & 0.516 & 0.573 & 0.365 \\
Finnish-ModernBERT-tiny-short & 0.217 & 0.211 & 0.533 & 0.598 & 0.390 \\
Finnish-ModernBERT-tiny-short-cpt & 0.218 & 0.214 & 0.517 & 0.595 & 0.386 \\
Finnish-ModernBERT-tiny-short-edu & 0.216 & 0.209 & 0.534 & 0.594 & 0.388 \\
mmBERT-base & 0.281 & 0.241 & 0.551 & 0.683 & 0.439 \\
XLM-RoBERTa-large & \textbf{0.290} & \textbf{0.248} & 0.546 & 0.583 & 0.417 \\
\bottomrule
\end{tabular}
\caption{Models' performances across the BEIR subset using the nDCG@10 metric. Results with "*" are reported by \protect{\citet{modernbert}}.  The best results for monolingual and multilingual models are highlighted in bold. }
\label{tab:performance-bier}
\end{table}

\end{document}